%% file: main.tex
\documentclass[runningheads]{llncs}
\usepackage{graphicx}
\usepackage{fancyhdr}

\usepackage{epsfig}
\usepackage{amsmath}
\usepackage{amssymb}

\usepackage{mathtools}
\usepackage{url}
\usepackage{enumitem}
\usepackage{subcaption}
\usepackage{multirow}

\usepackage{tikz}

\begin{document}
\title{PW-MAD: Pixel-wise Supervision for Generalized Face Morphing Attack Detection}
\titlerunning{PW-MAD: Pixel-wise Supervision for Generalized Face MAD}
%

\author{Naser Damer \inst{1,2} \and No\'{e}mie Spiller \inst{1} \and Meiling Fang \inst{1,2} \and Fadi Boutros \inst{1,2} \and \\ Florian Kirchbuchner \inst{1} \and Arjan Kuijper\inst{1,2}
}

\authorrunning{N. Damer et al.}

\institute{Fraunhofer Institute for Computer Graphics Research IGD, Germany \and
Department of Computer Science, TU Darmstadt, Germany
\email{naser.damer@igd.fraunhofer.de}\\}

\maketitle              

\begin{abstract}
A face morphing attack image can be verified to multiple identities, making this attack a major vulnerability to processes based on identity verification, such as border checks.
Various methods have been proposed to detect face morphing attacks, however, with low generalizability to unexpected post-morphing processes.
A major post-morphing process is the print and scan operation performed in many countries when issuing a passport or identity document.
In this work, we address this generalization problem by adapting a pixel-wise supervision approach where we train a network to classify each pixel of the image into an attack or not, rather than only having one label for the whole image. 
Our pixel-wise morphing attack detection (PW-MAD) solution proved to perform more accurately than a set of established baselines. 
More importantly, PW-MAD shows high generalizability in comparison to related works, when evaluated on unknown re-digitized attacks.
Additionally to our PW-MAD approach, we create a new face morphing attack dataset with digital and re-digitized samples, namely the LMA-DRD dataset that is publicly available for research purposes upon request.

\keywords{Face recognition  \and Face morphing \and Morphing attack detection.}
\end{abstract}
\section{Introduction}




The face recognition performance advances driven by deep-learning \cite{DBLP:conf/cvpr/DengGXZ19,boutros2021elasticface}, along with the relatively high social acceptance \cite{Jain:1998:BPI:552539}, have introduced face recognition technologies to security sensitive applications (e.g. ID/travel documents) \cite{FaceMarket2017}. However, face recognition systems are vulnerable to many attacks \cite{DBLP:journals/cviu/MassoliCAF21,DBLP:conf/bmvc/DamerD16,DBLP:conf/btas/DamerWBBT0K18}, one of these is face morphing attack.
Ferrara et al. \cite{DBLP:conf/icb/FerraraFM14} analyzed face morphing attacks early on by showing that one attack face image can be, automatically and by human experts, matched to more than one person. When morphing attacks are used in travel or identity documents, they allow multiple subjects to be verified to one document. This faulty subject link to the document identity can lead to a wide range of illegal activities, including financial transactions, illegal immigration, human trafficking, and circumventing criminal identity lists. 

Morphing attack detection solutions (MAD) are developed to classify an investigated face image into an attack or a bona fide (no attack). 
The performance of MAD solutions has been shown to drop substantially when facing unknown (not used in the training) variations in the investigated images. 
Such variations include the morphing technique \cite{MIPGAN,DBLP:conf/iwbf/VenkateshZRRDB20}, image compression \cite{DBLP:conf/visapp/MakrushinND17}, and re-digitization (print and scan) \cite{MADVgg} among others.
The re-digitization is one of the most studied of these variations as it represents the scenario where a printed image is presented when applying for an ID/travel document. This image is then scanned to be included on the document and possibly go through MAD. 
MAD solutions aiming at generalizability utilised handcrafted \cite{DBLP:conf/btas/RaghavendraRB16,DBLP:conf/cvpr/RaghavendraRVB17a} and deep learning features \cite{MADVgg,DBLP:conf/cvpr/RaghavendraRVB17a}, however, while using single binary target per image in the training.

This work aims at providing a generalizable MAD by proposing the adaption of pixel-wise supervision in the training process, giving the network the chance to distribute its focus on more evident manipulation effects, rather than, general image artifacts. This resulted in our pixel-wise supervised MAD (PW-MAD). To develop and evaluate our proposed solution, we additionally presented a publicly available morphing dataset that includes digital and re-digitized attacks and bona fide samples. 
Our solution proved to outperform a set of widely used baselines, especially when faced with unknown re-digitized images. Our PW-MAD has also shown better performance generalization over related works investigating the issue of re-digitized morphing attacks.

\section{Related work}

MAD methods can be separated into two main categories, single image and differential MAD \cite{survey}. Single image MAD only analyses the investigated image to build a decision of attack or bona fide \cite{DBLP:conf/icb/RaghavendraRVB17,DBLP:conf/cvpr/RaghavendraRVB17a,DBLP:conf/fusion/DamerZWSKK19,DBLP:conf/iwbf/AghdaieCSDN21}. The differential MAD uses the investigated image and a live image (assuming that the process allows for that). Differential MAD analyses the relation between both images to build a decision of attack or bona fide  \cite{DBLP:conf/dagm/DamerBWBTBK18,DBLP:journals/tifs/ScherhagRMB20,DBLP:conf/icb/DamerSZWTKK19,DBLP:conf/icpr/SoleymaniCDDN20}. This work focuses on single image MAD as it demands fewer requirements on the use-case.

Single image MAD solutions can be roughly categorized into ones using handcrafted features and ones using deep learning features. 
Such handcrafted features included Binarized Statistical Image Features (BSIF) \cite{DBLP:conf/iwbf/ScherhagRRGRB17,DBLP:conf/btas/RaghavendraRB16}, Local Binary Patterns (LBP) \cite{DBLP:conf/btas/DamerS0K18}, Local Phase Quantization (LPQ) \cite{DBLP:conf/icb/RaghavendraRVB17}, or features established in the image forensic analyses such as the photo response non-uniformity (PRNU) \cite{DBLP:conf/iciap/DebiasiDSRSBKU19}. The MAD solutions based on deep learning commonly used pre-trained networks with or without fine-tuning, such as versions of VGG \cite{MADVgg}, AlexNet \cite{DBLP:conf/cvpr/RaghavendraRVB17a}, or networks trained for face recognition purposes such as OpenFace \cite{DBLP:conf/fusion/DamerZWSKK19}. However, all these works, used a single binary label as the target of their training.

Many of these works have raised the issue of the generalizability of the MAD decisions when facing variabilities in the face morphing or image handling process. Such variabilities included the synthetic image generation processes \cite{DBLP:conf/btas/DamerGZKK19,DBLP:conf/btas/DamerBSKK19,MIPGAN,Regenmorph}, different data sources \cite{DBLP:conf/iwbf/ScherhagRB18}, morphing pair selection \cite{DBLP:conf/icb/DamerSZWTKK19},
image compression
\cite{DBLP:conf/visapp/MakrushinND17}, and re-digitization  \cite{MADVgg,DBLP:conf/btas/RaghavendraRB16,DBLP:conf/cvpr/RaghavendraRVB17a}. These variabilities have been shown to cause a drop in the MAD performance when they were unknown in the MAD training phase. 
The most practically relevant of these is the re-digitization, as it reflects the practice of requiring a printed image for travel/identity document issuance, where the authorities would scan this printed image. For this case, a number of private databases were created along with the analyses of the generalization of different MAD solutions such as different versions of VGG \cite{MADVgg}, AlexNet \cite{MADVgg}, the fusion of different pre-trained networks \cite{DBLP:conf/cvpr/RaghavendraRVB17a}, and the BSIF features with an SVM classifier \cite{DBLP:conf/btas/RaghavendraRB16,DBLP:conf/cvpr/RaghavendraRVB17a}. All these works have shown a substantial drop in the MAD performance on the re-digitized attacks when they were not used in the training process, which is the research gap addressed in this paper.


\section{Methodology}


This section presents our proposed PW-MAD approach along with a set of baseline approaches.

\subsection{The proposed PW-MAD}


Our proposed PW-MAD solution takes advantage of pixel-level supervision, i.e. a label of attack or bona fide for each image pixel, rather than being only supervised by one label for the whole image. This enhances the ability of the algorithm to distribute (spatially) its focus on areas with more evident manipulation. This is performed with the aim of bringing less focus on non-attack-related artifacts, and thus enhance the generalizability of the MAD decision. Such a supervision approach has been shown to gain these benefits when dealing with the generalizability of detecting iris and face spoofing attacks \cite{deeppix_19,DBLP:conf/icb/FangDBKK21}, however, it was never applied to the generalizability sensitive MAD. 

Our PW-MAD utilizes a densely connected network framework for MAD with binary and deep pixel-wise supervision. 
This framework is based on the DenseNet \cite{densenet} architecture, as motivated in \cite{deeppix_19}. Specifically, we use the DenseNet-121 architecture \cite{densenet}.
The use of this architecture is motivated by the high performances achieved in detecting iris and face spoofing attacks \cite{deeppix_19,DBLP:conf/icb/FangDBKK21}.
The architecture is modified to be simpler with only two dense blocks and two transition blocks with a fully connected layer with sigmoid activation to produce the binary output. 
In addition, a convolution layer with a kernel size of $1 \times 1$ is added before this fully connected layer, to generate the feature map for pixel-wise supervision. The feature map (size of $14 \time 14$ in our case) generated from this convolution layer is used to supervise the training of the network in a pixel-wise manner. Finally, the network is trained under pixel-wise and binary supervision. For the loss function, Binary Cross-Entropy (BCE) is used for both pixel-wise and binary supervision. The equation of BCE is:
\[
\mathcal{L}_{BCE} = -[y \cdot \log x + (1-y)\cdot \log (1-x)]
\]
where $y$ presents the ground truth label. $x$ is predicted probability. We use the $\mathcal{L}_{BCE}^{PW}$ to indicate the loss computed based on pixel-wise feature map and $\mathcal{L}_{BCE}^{B}$ is the loss computed based on binary output. Thus, an overall loss $\mathcal{L}_{overall}$ is formulated as 
\[
\mathcal{L}_{overall} = \lambda \cdot \mathcal{L}_{BCE}^{PW} + (1 - \lambda) \cdot \mathcal{L}_{BCE}^{B},
\] 
where $\lambda$ is set to 0.5 in the experiments.

Furthermore, we use the hyper-parameters (Adam optimizer with a learning rate of $10^{-4}$ and weight decay of $10^{-5}$) as motivated in \cite{deeppix_19} for the training. Additionally, we apply class weight and early stopping techniques to avoid overfitting. 
The final score for each test image is computed by binary output.

\subsection{Baselines}

\textbf{LBP:} The local binary patterns (LBP) are used extensively in MAD solutions with satisfactory results \cite{DBLP:conf/eusipco/SpreeuwersSV18,DBLP:conf/btas/DamerS0K18}. The face in a frame is first detected, cropped, and normalized into a size of $64 \times 64$ pixels. Then, an RGB face is converted into HSV and YCbCr color spaces. Third, the LBP features are extracted from each channel. The multi-channel extraction of LBP features has been shown to enhance the performance of MAD in a number of previous works \cite{DBLP:conf/icb/RaghavendraRVB17,DBLP:conf/isba/RamachandraVRB19}. The obtained six LBP feature vectors are then concatenated into one feature vector to feed into a Softmax classifier, resulting in a decision score.

\noindent \textbf{VGG16:} The VGG16 architecture \cite{DBLP:journals/corr/SimonyanZ14a} is used extensively in MAD solutions with very competitive results \cite{MADVgg,DBLP:conf/cvpr/RaghavendraRVB17a}. 
The used network is pretrained on large-scale ImageNet dataset \cite{DBLP:conf/cvpr/DengDSLL009} and provided as a pretrained network in \cite{DBLP:journals/corr/SimonyanZ14a}.
Before processing the image, it is normalized to $224 \times 224$ pixels, then extracts the output of an intermediate layer of VGG16 which is used as a feature. The features are scaled before they are fed to a linear SVM classifier.

\noindent $\mathrm{\mathbf{Inception_{FT}}}$ and $\mathrm{\mathbf{Inception_{TFS}}}$: This baseline uses the Inception-v3 \cite{inception_v3} network architecture as the cornerstone. This architecture has been used successfully for MAD \cite{DBLP:conf/cvip/RamachandraVRB18} and fake face detection approaches \cite{DBLP:conf/biosig/KhodabakhshRRWB18}. 
We report the results of a fine-tuned version of the pre-trained Inception-v3, this will be referred to as $\mathrm{Inception_{FT}}$ (the pre-trained network is trained on ImageNet dataset \cite{DBLP:conf/cvpr/DengDSLL009} and made available by \cite{inception_v3}). The last classification layer of Inception-v3 is modified to fit our two classes case where an input image is either bona fide or attack. Only the weight of this classification layer is fine-tuned, while the weights of other layers are fixed. 
We also report the results of a trained from scratch Inception-v3 model, named $\mathrm{Inception_{TFS}}$. 
In the training phase, the binary cross-entropy loss function and Adam optimizer with a learning rate of $10^{-4}$ and a weight decay of $10^{-5}$ are used. Moreover, the early stopping techniques used in our PW-MAD method are also applied for training of $\mathrm{Inception_{FT}}$ and $\mathrm{Inception_{TFS}}$ to avoid overfitting and for a fair comparison.

It must be noted that our PW-MAD and all the baseline solutions used only the training part of the data for training (and the development split for validation when training a neural network). All the three splits, train, development, and test, are identity-disjoint. 

\section{Experimental setup}

This section presents our newly created morphing attack dataset (with vulnerability analyses) along with the experimental settings and evaluation metrics.
\subsection{The dataset}
\label{sec:exp:db}

As there is no suitable publicly available morphing dataset, we opted to create a carefully designed morphing dataset that is described in this section. This dataset will be referred to as the digital and re-digitized landmark-based morph dataset (LMA-DRD). The dataset is built on the VGGFace-2 dataset \cite{Cao.2017}, which is composed of 3.31 million images of 9131 identities. This basic dataset was chosen as it has a large number of images per subject, which allows the choice of high-quality samples as will be explained in this section. The images are not scaled and therefore have different resolutions, however as will be clarified, we chose high-resolution images. To cover the frontal image condition in the International Civil Aviation Organisation (ICAO) travel document requirements \cite{ICAO}, all non-frontal images are filtered out by detecting the central coordinate of the eyes and the upper coordinate of the nose. The two distances between each of the two eyes and the nose landmarks are calculated, and if the ratio of the difference between these distances to any of them was more than $0.05$, the image is neglected. The detected landmarks were used to ensure that all the considered images had an eye-to-eye distance of at least 90 pixels as defined in \cite{ICAO}. Based on these criteria, the total number of images after filtering was 54010 images. This cleaned version of the data is the one that all the samples in our LMA-DRD dataset originate from.

\begin{figure}[h!]
    \centering
    \includegraphics[width=0.75\textwidth]{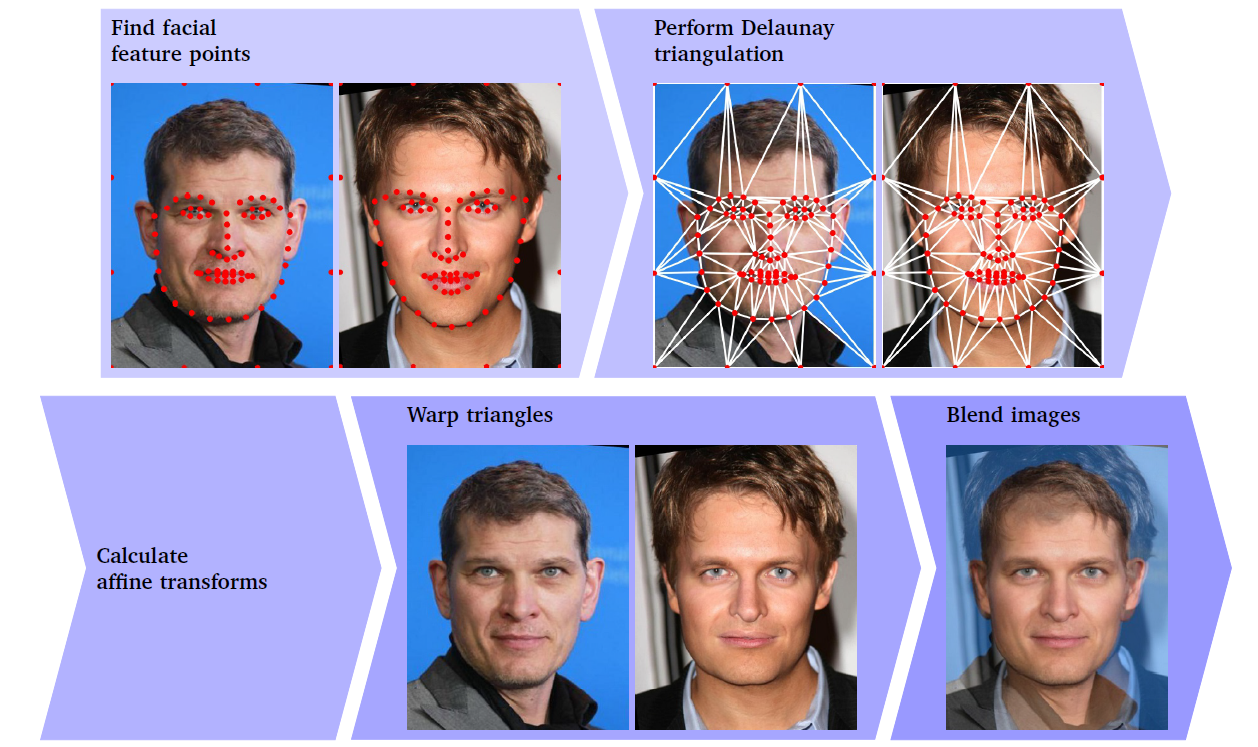}
    \caption{Face Morphing pipeline followed in this work and described in \cite{DBLP:conf/icb/RaghavendraRVB17}. The pipeline starts with detecting landmarks in the original (to be morphed) images, and ends up with the blended morphed image.}
    \label{fig:morph_pipeline}
\end{figure}

From the filtered data, and as a starting point, 197 images of 197 identities were manually chosen so that they are split evenly between males and females, frontal faces, with a neutral expression, have no glasses, good illumination quality, and no occlusion. 
Each of these images is was paired with the two most similar faces of two different images of different identities.
This is the typically recommended protocol, which makes sense if the goal is to create a confident attack \cite{DBLP:conf/icb/DamerSZWTKK19}. 
This pairing depended on the similarity between the key image and the selected paired images. The similarity was measured by the Euclidean distance between the OpenFace representations \cite{amos2016openface}. 
The 197 key images were paired twice, resulting in 394 morphing pairs. The paired images (besides the key ones) were not paired with more than one key image. For each of these pairs, a morphed image has been created using the landmark-based approach and parameters presented in \cite{DBLP:conf/icb/RaghavendraRVB17}, the morphing pipeline is illustrated in Figure \Ref{fig:morph_pipeline}.
The 394 morphed images were manually inspected and any image with strong artifacts was removed, this resulted in a final 276 morphing attack images. From the same identities involved in the attacks, a second bona fide image was chosen (total 591 images), these were manually filtered for quality as described above to comply with ICAO standards. After filtering, the remaining images were 364 images. These 364 images are considered bona fide samples.
In total, the created morphing LMA-DRD dataset contains 364 digital Bona fide (D-BF) images and 276 digital morphing attacks (D-M). These images were printed on 11,5cmx9cm glossy photo paper in a professional studio and scanned with 600dpi scanner. 
They resulted in the same number of re-digitized bona fide (PS-BF) and attacks (PS-M).
The resulted LMA-DRD dataset is split into three identity-disjoint parts, train, development, and test, splits. The splits are done so that they are identity-disjoint, have a similar number of samples, and are equally distributed over males and females to the best possible extend.
In our experiments, the training uses only the training data split, the validation during the training (when training a neural network) uses only the development set, and the evaluation is performed only on the test set. 
The vulnerability analyses in the next paragraph are performed on the the three data splits (train, develop, and test) as the analyses do not include any training.
Table \ref{tab:data} present an overview of our LMA-DRD dataset and its splits.
Samples of the images included in the dataset and morphing results are shown in Fig. \ref{fig:samp}.

\input{samples_figure}
\input{data_table}

A ResNet-100 ArcFace \cite{DBLP:conf/cvpr/DengGXZ19} pre-trained face recognition model is used to analyse the vulnerability of face recognition to the presented attacks, as it is one of the most widely used and best performing academic face recognition models with a publicly available pre-trained network.
The vulnerability is measured as the Mated Morph Presentation Match Rate (MMPMR) (as defined in \cite{DBLP:conf/biosig/ScherhagNRGVSSM17}) and is presented in Table \ref{tab:vul} for a false match rate (FMR) of 0.1\% (as recommended for border check operations by Frontex \cite{frontex2015best}) and 1.0\%, which proves the validity of the considered attacks. 
An MMPMR of 91.30\% at FMR of 1.0\% means that 91.30\% of the attacks will be matched to both contributing identities if the considered face recognition solution uses the decision threshold at 1.0\% FMR.
We notice, in table \ref{tab:vul}, that the vulnerability to re-digitized attacks is slightly less than it is to digital attacks. This might be due to the image artifacts introduced in the re-digitization process.
We additionally provide a visual illustration of the face recognition vulnerability to the attacks in Figure \ref{fig:vul}. The figures plot the similarity score between the attacks (M-D in \ref{fig:vul}.a and M-PS in \ref{fig:vul}.b) and the first involved identity (x-axis) vs. the one with the second identity (y-axis). The red lines in these plots represent the threshold value that achieves FMR of 0.1\%.
This helps to put the plotted scores in perspective knowing that any attack represented by a dot in the figure successfully match both identities at this threshold (FMR = 0.1\%) if it is above and to the right of the red lines. The plots in Figure \ref{fig:vul}, confirm the MMPMR values in Table \ref{tab:vul} by showing the high vulnerability of the face recognition system to the presented attacks and the slight drop in this vulnerability after the re-digitization process.
The LMA-DRD data is publicly available to researchers upon request. 

\subsection{Experiments and evaluation metrics}

\input{vul_table2}
\input{scatterplot}

Our experiments aim at evaluating the generalizability of our proposed PW-MAD and the other baseline MADs.
As baseline experiments, we evaluate the different MADs on the same type of data (digital or re-digitized).
This results in two baseline experimental settings, one uses the digital data for training and testing (Train-D Test-D) and one uses the re-digitized data for training and testing (Train-PS Test-PS). 
Two additional experimental setups measure the generalizability on data of an unknown type.
One uses the digital data for training and re-digitized data for testing (Train-D Test-PS) and one uses the re-digitized data for training and digital data for testing (Train-PS Test-D).  
It must be noted that the "Train-D Test-PS" reflects the most application-relevant use-case and thus the most commonly reported case on MAD generalization in the literature \cite{DBLP:conf/btas/RaghavendraRB16,DBLP:conf/cvpr/RaghavendraRVB17a,MADVgg}.
In our experiments, the training uses only the training data split, the validation during the training (when training a neural network) uses only the development set, and the evaluation is performed only on the test set. The three sets are identity-disjoint to prevent biases in the evaluation.


The MAD performance is presented by the Attack Presentation Classification Error Rate (APCER), i.e. the proportion of attack images incorrectly classified as bona fide samples, and the Bona fide Presentation Classification Error Rate (BPCER), i.e. the proportion of bona fide images incorrectly classified as attack samples, as defined in the ISO/IEC 30107-3 \cite{ISO301073}. To cover different operation points, and to present the comparative results, we report the BPCER at three different fixed APCER values (0.1\%, 1.0\%, and 10.0\%). 
To provide a visual evaluation on a wider operation range, we plot receiver operating characteristic (ROC) curves by plotting the APCER on the x-axis and 1-BPCER on the y-axis at different operational points.
It must be noted again that the MAD evaluation was performed only on the identity-disjoint test data as described in Section \ref{sec:exp:db}.

\section{Results and discussion}

\begin{figure}[h!]
    \centering
    \includegraphics[width=0.99\textwidth]{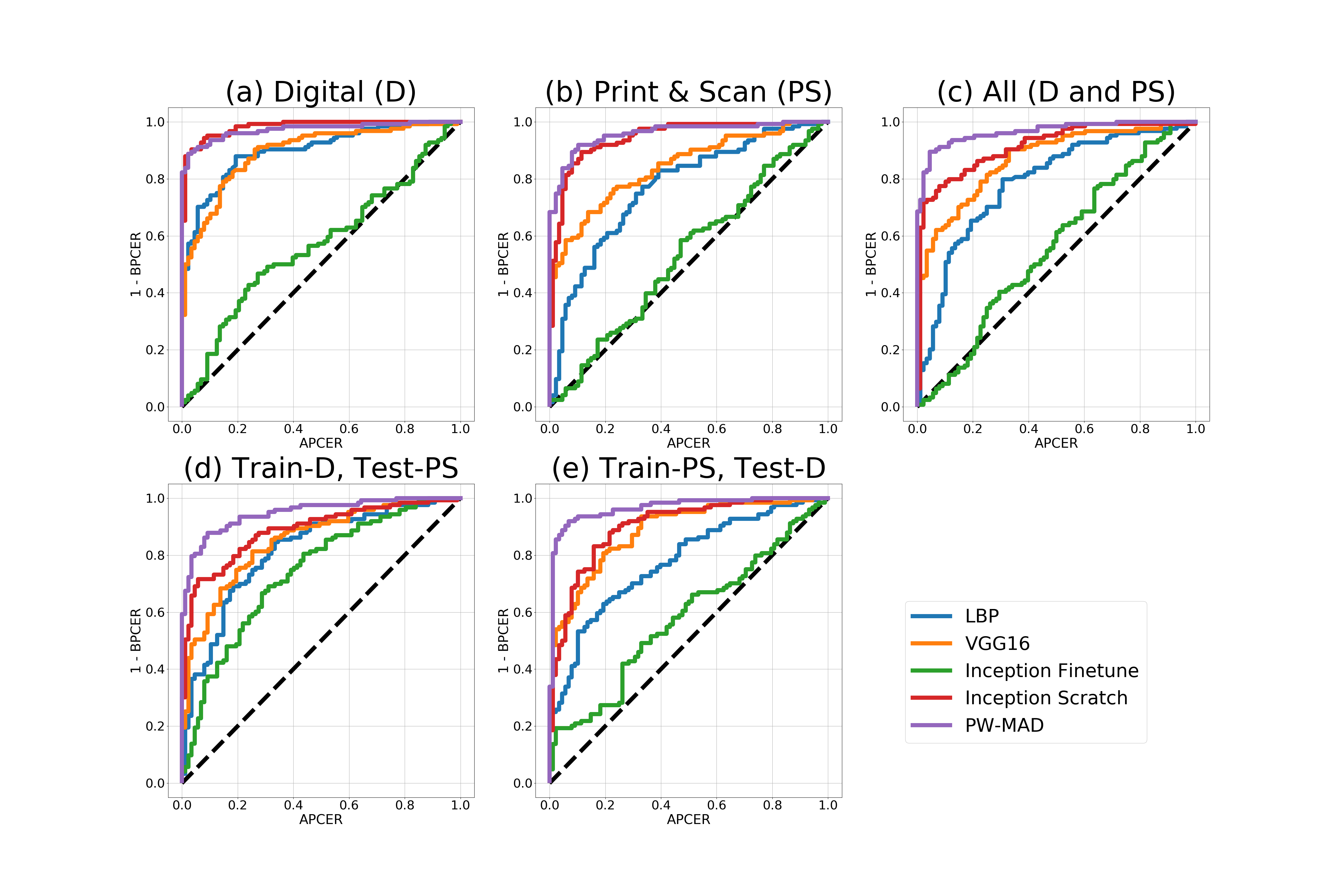}
    \caption{ROC curves achieved by our PW-MAD solution along with the other baselines. The considered experimental settings are Train-D Test-D in (a), Train-PS Test-PS in (b), Train-D and PS, Test-D and PS in (c), Train-D Test-P in (d), and Train-PS Test-D in (e). Note the superior performance of the PW-MAD, especially in the inter-data type settings in (d) and (e). }
    \label{fig:roc}
\end{figure}

Table \ref{tab:res} lists the BPCER rates achieved at different APCER thresholds for the PW-MAD and the baseline MADs. 
On the intra-data type settings (Train D Test D) the PW-MAD solution outperforms all baselines at the lowest APCER operation point (0.1\%) by scoring a BPCER of 17.74\% in comparison to 34.67\% for the next best MAD. For higher APCER values, the PW-MAD scores the second-best BPCER.
Also for the intra-data type settings (Train PS Test PS), the PW-MAD scores the lowest BPCER (best) on all operational points (APCER thresholds). This is supported by the ROC curves in Figures \ref{fig:roc}.a and \ref{fig:roc}.b.

In the more challenging inter-data type settings, the proposed PW-MAD outperformed all the baselines at all the APCER thresholds.
For the "Train-D Test-PS" setting, the BPCER (at APCER of 1.0\%) scored by our PW-MAD is 32.52\% in comparison to 49.59\% for the next best MAD. This constitutes a 34.4\% drop in the BPCER value.
For the "Train-PS Test-D" setting, the BPCER (at APCER of 1.0\%) scored by our PW-MAD is 19.35\% in comparison to 51.61\% for the next best MAD. This constitutes a 62.5\% drop in the BPCER value.
These inter-data type evaluation results demonstrate the superior generalizability of our proposed PW-MAD in comparison to the baselines. These inter-data type conclusions are supported by the ROC curves in Figures \ref{fig:roc}.d and \ref{fig:roc}.e, where the better maintenance of the performance (in comparison to the baselines) is apparent when comparing these curves to the ones in Figures \ref{fig:roc}.a and \ref{fig:roc}.b.

\input{Reaults_table}

To put the generalizability of the proposed approach in perspective, we compare our PW-MAD with the previously published approaches that targeted the detection of re-digitized morphing attacks \cite{DBLP:conf/cvpr/RaghavendraRVB17a,MADVgg,DBLP:conf/icb/RaghavendraRVB17}. As these works reported their results on private datasets, we are not able to build a direct performance comparison. We rather present the reported results when the MAD is trained and tested on digital morphs, along with the performance of the MAD trained on digital morphs and tested on re-digitized morphs, see Table \ref{tab:comp_sota} where the performances are reported in BPCER at APCER of 10\% as it is the common reported measure between the relevant previous works. We also list the BPCER error increase (in percentage points) when moving from testing on the known digital morphs to the unknown re-digitized. This performance drop might not be an optimal measure of the performance, as it neglects the absolute performance, but it rather gives a clear indication of the generalization. It is noted in Table \ref{tab:comp_sota} that our proposed PW-MAD results in the lowest performance drop between previously reported results, indicating the relatively high generalizability of its decisions.
When it comes to training on re-digitized attacks and testing on digital attacks, our PW-MAD actually gains performance, BPCER at 10\% APCER moves from 8.13\% to 6.45\% as in Table \ref{tab:res}. This training/testing protocol was only reported in previous literature in \cite{DBLP:conf/cvpr/RaghavendraRVB17a}, where their transfer learning approach reported in the best-case scenario, a BPCER at 10\% APCER of 16.43\% on known re-digitized attacks and dropping to 30.13\% when testing on the unknown digital attacks. This again points out the relative generalizability of our proposed PW-MAD approach. 

\input{comparison_table}

\section{Conclusion}

This work targeted the enhancement of the generalizability of MAD performance.
This is achieved by proposing the PW-MAD solution that leverages the adaption of pixel-wise supervision into the training process to produce a stable performance, even when facing unknown variations like re-digitized images.
We presented a new dataset that included digital and re-digitized samples, allowing the development and evaluation of the proposed PW-MAD.
The PW-MAD proved to provide a superior MAD generalizability over a set of widely used baselines and previously reported results in state-of-the-art.

\paragraph{Acknowledgment:} This research work has been funded by the German Federal Ministry of Education and Research and the Hessian Ministry of Higher Education, Research, Science and the Arts within their joint support of the National Research Center for Applied Cybersecurity ATHENE.

%
%
%

 \bibliographystyle{splncs04}
 \bibliography{main}
\end{document}

%% file: samples_figure.tex
\begin{figure}[h!]
    \centering
    \begin{subfigure}[b]{0.16\textwidth}
         \centering
         \includegraphics[width=0.99\textwidth]{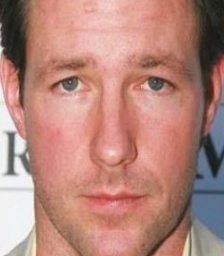}
         \includegraphics[width=0.99\textwidth]{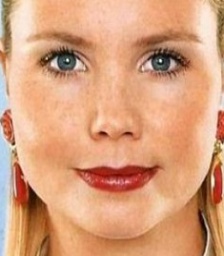}
         \caption{\tiny{D-BF - ID1}}
         \label{fig:samp:d_id1}
     \end{subfigure}
     \begin{subfigure}[b]{0.16\textwidth}
         \centering
         \includegraphics[width=0.99\textwidth]{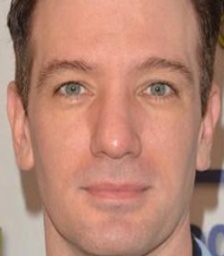}
         \includegraphics[width=0.99\textwidth]{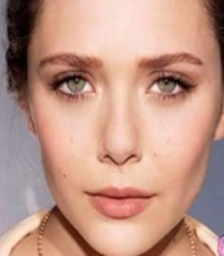}
         \caption{\tiny{D-BF - ID2}}
         \label{fig:samp:d_id2}
     \end{subfigure}
     \begin{subfigure}[b]{0.16\textwidth}
         \centering
         \includegraphics[width=0.99\textwidth]{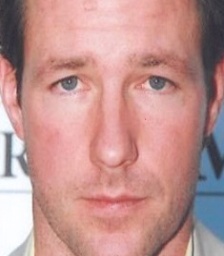}
         \includegraphics[width=0.99\textwidth]{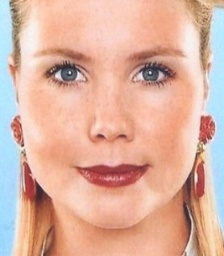}
         \caption{\tiny{PS-BF - ID1}}
         \label{fig:samp:PS_id1}
     \end{subfigure}
     \begin{subfigure}[b]{0.16\textwidth}
         \centering
         \includegraphics[width=0.99\textwidth]{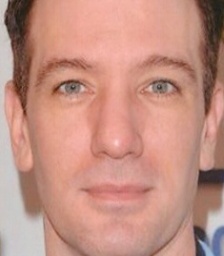}
         \includegraphics[width=0.99\textwidth]{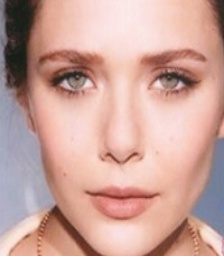}
         \caption{\tiny{PS-BF - ID2}}
         \label{fig:samp:PS_id2}
     \end{subfigure}
     \begin{subfigure}[b]{0.16\textwidth}
         \centering
         \includegraphics[width=0.99\textwidth]{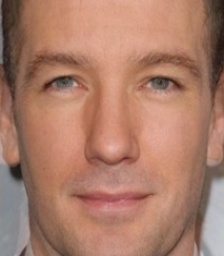}
         \includegraphics[width=0.99\textwidth]{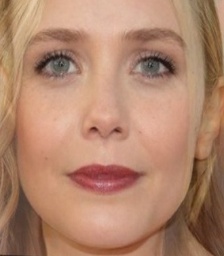}
         \caption{\tiny{D-M}}
         \label{fig:samp:D_M}
     \end{subfigure}
     \begin{subfigure}[b]{0.16\textwidth}
         \centering
         \includegraphics[width=0.99\textwidth]{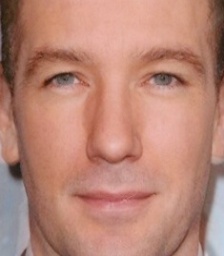}
         \includegraphics[width=0.99\textwidth]{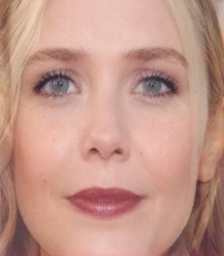}
         \caption{\tiny{PS-M}}
         \label{fig:samp:PS_M}
     \end{subfigure}
     \caption{Samples of our LMA-DRD database with the digital bona fide samples ((a) and (b)), re-digitized bona fide ((c) and (d)), digital morphing attacks (e), and re-digitized morphing attacks (f).  }
    \label{fig:samp}
\end{figure}

%% file: data_table.tex
\begin{table}[h!]
\footnotesize
\centering
\begin{tabular}{|l|l|l|l|l|}
\hline
            & \multicolumn{2}{c|}{Attacks}                                                  & \multicolumn{2}{c|}{Bona fide}                                                  \\ \hline
            & \multicolumn{1}{c|}{Digital (D-M)} & \multicolumn{1}{c|}{Re-digitized (PS-M)} & \multicolumn{1}{c|}{Digital (D-BF)} & \multicolumn{1}{c|}{Re-digitized (PS-BF)} \\ \hline
Train       & 96                                  & 96                                        & 121                                & 121                                      \\ \hline
Development & 92                                  & 92                                        & 120                                & 120                                      \\ \hline
Test        & 88                                  & 88                                        & 123                                & 123                                      \\ \hline \hline
Total       & 276                                 & 276                                       & 364                                & 364                                      \\ \hline
\end{tabular}
\caption{A detailed view of the presented LMA-DRD database. The numbers indicate the number of images in each data type and data split. Note that the training, development, and testing splits of the data are all identity-disjoint. }
\label{tab:data}
\end{table}

%% file: vul_table2.tex
\begin{table}[h!]
\centering
\begin{tabular}{|l|l|l|}
\hline
\multicolumn{1}{|c|}{\multirow{2}{*}{Attack}} & \multicolumn{1}{c|}{at FMR=0.1\%} & \multicolumn{1}{c|}{at FMR=1\%} \\ \cline{2-3} 
\multicolumn{1}{|c|}{}                        & \multicolumn{1}{c|}{MMPMR(\%)}    & \multicolumn{1}{c|}{MMPMR(\%)}  \\ \hline
D-M                                           & 80.07                             & 91.30                           \\ \hline
PS-M                                          & 77.17                             & 88.41                           \\ \hline
\end{tabular}
\caption{The vulnerability to the LMA-DRD dataset attacks, both the digital (D-M) and the re-digitized (PS-M) represented by the MMPMR(\%) at two different decision thresholds (FMR=0.1\% and 1.0\%) of the investigated ResNet-100 Arcface pre-trained model. Note the slight decrease in vulnerability to the re-digitized attacks when compared to the digital ones. }
\label{tab:vul}
\end{table}

%% file: scatterplot.tex
\begin{figure}[h!]
    \centering
    \begin{subfigure}[b]{0.49\textwidth}
         \centering
         \includegraphics[width=1.0\textwidth]{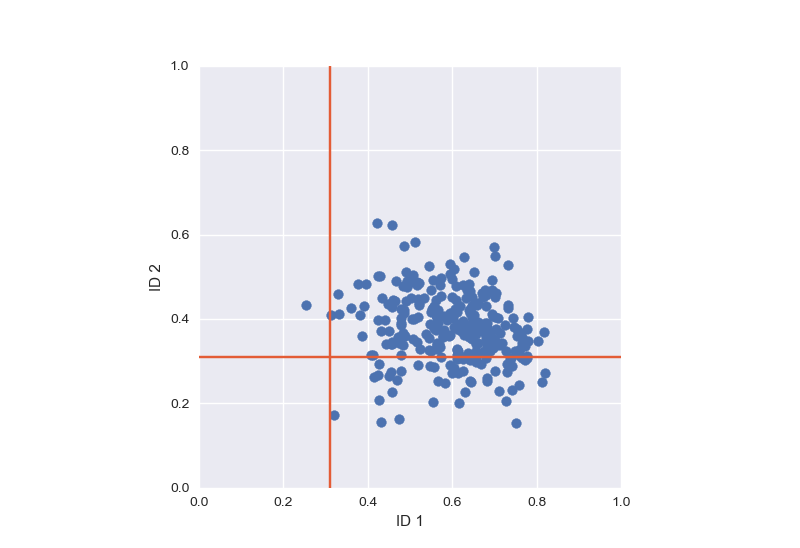}
         \caption{M-D}
         \label{fig:sc:d}
     \end{subfigure}
     \begin{subfigure}[b]{0.49\textwidth}
         \centering
         \includegraphics[width=1.0\textwidth]{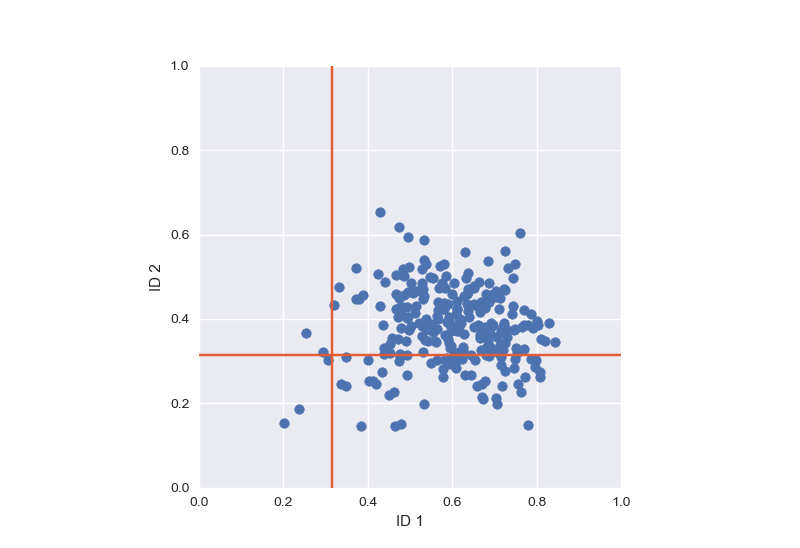}
         \caption{M-PS}
         \label{fig:sc:ps}
     \end{subfigure}
     \caption{The vulnerability of face recognition to the attacks in the LMA-DRD dataset attacks is represented by the similarity of the attack to the two identities used to create the attack (ID1 on the X-axis and ID2 on the y-axis). The red lines represent the similarity threshold for the FMR of 0.1\%, which indicates that all the attacks on the top and to the right of the red lines do match the targeted identities at this FMR setting. 
     Plot (a) represents the digital attacks, and (b) represents the re-digitized attacks.}
    \label{fig:vul}
\end{figure}

%% file: Reaults_table.tex
\begin{table}[h!]
\centering
\resizebox{\linewidth}{!}{%
\begin{tabular}{|l|l|l|l|l|l||l|l|l|l|l|l|}
\hline
\multicolumn{1}{|c|}{\multirow{2}{*}{Approach}} & \multicolumn{1}{c|}{\multirow{2}{*}{Train}} & \multicolumn{1}{c|}{\multirow{2}{*}{Test}} & \multicolumn{3}{c||}{BPCER (\%) @ APCER =} & \multicolumn{1}{c|}{\multirow{2}{*}{Approach}} & \multicolumn{1}{c|}{\multirow{2}{*}{Train}} & \multicolumn{1}{c|}{\multirow{2}{*}{Test}} & \multicolumn{3}{c|}{BPCER (\%) @ APCER =} \\ \cline{4-6} \cline{10-12} 
\multicolumn{1}{|c|}{}                          & \multicolumn{1}{c|}{}                       & \multicolumn{1}{c|}{}                      & 0.1\%          & 1.0\%          & 10.0\%        & \multicolumn{1}{c|}{}                          & \multicolumn{1}{c|}{}                       & \multicolumn{1}{c|}{}                      & 0.1\%        & 1\%          & 10\%        \\ \hline
LBP                                             & D                                           & D                                          & 51.61        & 51.61        & 25.80       & LBP                                            & PS                                          & PS                                         & 98.37        & 95.93        & 57.72       \\ \hline
VGG16                                           & D                                           & D                                          & 67.74        & 50.00        & 32.25       & VGG16                                          & PS                                          & PS                                         & 63.41        & 54.47        & 39.83       \\ \hline
Incep.$_{FT}$                                      & D                                           & D                                          & 98.38        & 97.58        & 81.45       & Incep.$_{FT}$                                     & PS                                          & PS                                         & 97.56        & 97.56        & 91.05       \\ \hline
Incep.$_{TFS}$                                      & D                                           & D                                          & 34.67        & \textbf{12.09}        & \textbf{4.83}        & Incep.$_{TFS}$                                     & PS                                          & PS                                         & 71.54        & 48.78        & 13.00       \\ \hline
PW-MAD                                          & D                                           & D                                          & \textbf{17.74}        & 16.12        & 6.45        & PW-MAD                                         & PS                                          & PS                                         & \textbf{31.70}        & \textbf{31.70}        & \textbf{8.13}        \\ \hline \hline
LBP                                             & D                                           & PS                                         & 96.74        & 80.48        & 51.21       & LBP                                            & PS                                          & D                                          & 75.80        & 75.00        & 46.77       \\ \hline
VGG16                                           & D                                           & PS                                         & 80.48        & 74.79        & 40.65       & VGG16                                          & PS                                          & D                                          & 70.96        & 51.61        & 33.06       \\ \hline
Incep.$_{FT}$                                      & D                                           & PS                                         & 95.93        & 94.30        & 62.60       & Incep.$_{FT}$                                     & PS                                          & D                                          & 95.16        & 86.29        & 79.03       \\ \hline
Incep.$_{TFS}$                                      & D                                           & PS                                         & 69.91        & 49.59        & 28.45       & Incep.$_{TFS}$                                     & PS                                          & D                                          & 81.45        & 62.09        & 25.80       \\ \hline
PW-MAD                                          & D                                           & PS                                         & \textbf{40.65}        & \textbf{32.52}        & \textbf{12.19}       & PW-MAD                                         & PS                                          & D                                          & \textbf{66.12}        & \textbf{19.35}        & \textbf{6.45}        \\ \hline
\end{tabular}
}
\caption{The BPCER at different APCER values (the lower BPCER, the better the MAD performance) achieved by our PW-MAD and the different considered baselines. One can note the better performance of the proposed PW-MAD on most experimental settings, especially when considering the realistic cross-attack scenario on the bottom two tables. Train and Test indicate the data type used for training and testing (digital (D) or re-digitized (PS)). The lowest BPCER for each train/test setup is in bold for each APCER threshold.}
\label{tab:res}
\end{table}

%% file: comparison_table.tex
\begin{table}[h!]
\centering
\resizebox{\linewidth}{!}{%
\begin{tabular}{|l|l|l|l|}
\hline
\multicolumn{1}{|c|}{}                           & \multicolumn{2}{c|}{BPCER (\%) @ APCER =   10\%} & \multicolumn{1}{c|}{}                                                                                             \\ \cline{2-3}
\multicolumn{1}{|c|}{\multirow{-2}{*}{Approach}} & Train-D, Test-D        & Train-D, Test-PS        & \multicolumn{1}{c|}{\multirow{-2}{*}{\begin{tabular}[c]{@{}c@{}}BPCER increase in \\ percentage points\end{tabular}}} \\ \hline
BSIF-SVM (P1) \cite{DBLP:conf/btas/RaghavendraRB16,DBLP:conf/cvpr/RaghavendraRVB17a}                                   & 38.25                  & 48.63                   & 10.38                                                                                                             \\ \hline
BSIF-SVM   (P2) \cite{DBLP:conf/btas/RaghavendraRB16,DBLP:conf/cvpr/RaghavendraRVB17a}                                   & 38.25                  & 57.53                   & 19.28                                                                                                             \\ \hline
Transferable   D-CNN (P1) \cite{DBLP:conf/cvpr/RaghavendraRVB17a}                        & 7.53                   & 24.65                   & 17.12                                                                                                             \\ \hline
Transferable   D-CNN (P2) \cite{DBLP:conf/cvpr/RaghavendraRVB17a}                       & 7.53                   & 17.8                    & 10.27                                                                                                             \\ \hline
Fine-tune   AlexNet \cite{MADVgg}                             & 0.8                    & 50.8                    & 50                                                                                                                \\ \hline
Fine-tune   VGG19 \cite{MADVgg}      & 0.8                    & 32.7                    & 31.9                                                                                                              \\ \hline
Fine-tune   VGG-Face16   \cite{MADVgg}                        & 0.8                    & 13.8                    & 13                                                                                                                \\ \hline
Fine-tune   VGG-Face2  \cite{MADVgg}                          & 0.0                    & 20                      & 20                                                                                                                \\ \hline
PW-MAD (ours)                                    & 6.45                   & 12.19                   & \textbf{5.74}                                                                                                              \\ \hline
\end{tabular}
} 
\caption{A comparison on the results presented in the state-of-the-art works reporting on experimental settings where the MAD is trained and tested on digital attacks (Tran-D, Test-D) and when trained on digital attacks and tested on re-digitized attacks (Train-D, Test-PS). The BPCER values are not directly comparable, as each of the works considered a different (private) database. 
The increase in BPCER percentage points represents the generalizability of the MAD to unknown variations in the attack and it shows that our proposed PW-MAD achieves the lowest drop in the performance, and thus relatively high generalizability.
The lowest increase (percentage points) in BPCER error between the two experimental setups is in bold. P1 and P2 indicate using different printers in the respective papers.}
\label{tab:comp_sota}
\end{table}